\documentclass[12pt,a4paper]{article}

\usepackage[latin1]{inputenc}
\usepackage{latexsym,enumerate,bm,bbm,amsmath,amssymb}
\usepackage[pdftex]{graphicx,color}
\usepackage{natbib}

\begin{document}

\title{Supervised functional classification: A theoretical remark and some
comparisons}
\author{ Amparo Ba\'{\i}llo\footnote{Corresponding author. Phone: +34 914978640,
e-mail: amparo.baillo@uam.es}
\quad and \quad Antonio Cuevas\thanks{The research of both authors was partially supported
by Spanish grant MTM2007-66632 and the IV PRICIT program titled {\em Modelizaci\'on Matem\'atica y Simulaci\'on Num\'erica en Ciencia y
Tecnolog\'{\i}a} (SIMUMAT).}
\\
\footnotesize Departamento de An\'alisis Econ\'omico: Econom\'{\i}a
Cuantitativa, Univ. Aut\'onoma de Madrid, Spain\\
\footnotesize Departamento de Matem\'aticas, Univ. Aut\'onoma de Madrid, Spain}

\date{}
\maketitle

\begin{abstract}
The problem of supervised classification (or discrimination) with functional data is
considered, with a special interest on the popular $k$-nearest neighbors ($k$-NN) classifier.

First, relying on a recent result by C\'erou and Guyader (2006), we prove the
consistency of the $k$-NN classifier for functional data whose distribution
belongs to a broad family of Gaussian processes with triangular covariance functions.

Second, on a more practical side, we check the behavior of the $k$-NN method when
compared with a few other functional classifiers. This is carried out through
a small simulation study and the analysis of several real functional data sets.
While no global ``uniform'' winner emerges from such comparisons, the overall performance of
the $k$-NN method, together with its sound intuitive motivation and relative simplicity,
suggests that it could represent a reasonable benchmark for the classification
problem with functional data.

\

\noindent
\it Key words and phrases\rm. Supervised classification, functional data, projections method,
nearest neighbors, discriminant analysis.

\noindent
\it AMS 2000 subject classification\rm. Primary 62G07; secondary 62G20.
\end{abstract}

\newpage

\noindent
\bf 1. Introduction\rm

\

\noindent
\it 1.1 Some background on supervised classification\rm

\

Supervised classification is the modern name for one of the oldest statistical problems in
experimental science: to decide whether an individual, from which just a random
measurement $X$ (with values in a ``feature space'' ${\cal F}$ endowed with a metric $D$)
is known, either belongs to the population $P_0$ or to $P_1$.
For example, in a medical problem $P_0$ and $P_1$ could correspond to the group of ``healthy''
and ``ill'' individuals, respectively. The decision must be taken from the
information provided by a ``training sample'' $\mathcal X_n = \{ (X_i,Y_i), 1\leq i\leq n \}$,
where $X_i$, $i=1,\ldots,n$, are independent replications of $X$, measured on $n$ randomly chosen
individuals, and $Y_i$ are the corresponding values of an
indicator variable which takes values 0 or 1 according to
the membership of the $i$-th individual to $P_0$ or $P_1$.
Thus the mathematical problem is to find a ``classifier''
$g_n(x)=g_n(x;\mathcal X_n)$, with $g_n:{\cal F}\rightarrow
\{0,1\}$, that minimizes the classification  error $P\{g_n(X)\neq Y\}$.

The term ``supervised'' refers to the fact that the individuals in the training
sample are supposed to be correctly classified, typically using ``external'' non
statistical procedures, so that they provide a reliable basis for the assignation
of the new observation. This problem, also known as ``statistical
discrimination'' or ``pattern recognition'', is at least 70 years old.
The origin goes back to the classical work by Fisher (1936) where, in the
$d$-variate case ${\cal F}={\mathbb R}^d$, a simple ``linear classifier''
$g_n(x)={\mathbbm 1}_{\{x: w^\prime x+w_0>0\}}$ was introduced (${\mathbbm 1}_A$ stands for
the indicator function of a set $A\subset {\cal F}$).

A deep insightful perspective of the supervised classification problem can be found in
the book of Devroye et al (1996). Other useful textbooks are Hand (1997) and Hastie et al.
(2001). All of them focus on the standard multivariate case ${\cal F}={\mathbb R}^d$.

It is not difficult to prove (e.g., Devroye et al., 1996, p. 11) that the optimal classification rule
(often called ``Bayes rule'') is
\begin{equation} \label{opt}
g^*(x)={\mathbbm 1}_{\{\eta(x)>1/2\}},
\end{equation}
where $\eta(x)=E(Y|X=x)$. Of course, since $\eta$ is unknown the exact
expression of this rule is usually unknown, and thus
different procedures have been proposed in order to
approximate it. In particular, it can be seen that
Fisher's linear rule is optimal provided that the conditional distributions of $X|Y=0$ and
$X|Y=1$ are both normal with identical covariance matrix. While these conditions look quite restrictive,
and it is straightforward
to construct problems where any linear rule has a
poor performance,
Fisher's classifier is still by far the most popular choice among users.

A simple non-parametric alternative is given by the $k$-nearest neighbors ($k$-NN)
method which is obtained by replacing the unknown regression function $\eta(x)$ in
(\ref{opt}) with the regression estimator
\begin{equation} \label{RegEstkNN}
\eta_n(x) = \frac{1}{k} \sum_{i=1}^n {\mathbbm 1}_{\{ X_i\in k(x) \}} Y_i
\end{equation}
where $k=k_n$ is a given (integer) smoothing parameter and ``$X_i\in k(x)$'' means that
$X_i$ is one of the $k$ nearest neighbours of $x$.
More concretely, if the pairs $(X_i,Y_i)_{1\leq i \leq n}$ are re-indexed as
$(X_{(i)},Y_{(i)})_{1\leq i \leq n}$ so that the $X_{(i)}$'s are arranged
in increasing distance from $x$,
$D(x,X_{(1)}) \leq D(x,X_{(2)}) \leq \ldots \leq D(x,X_{(n)})$,
then $k(x) = \{ X_{(i)},1\leq i \leq k \}$.
This leads to the $k$-NN classifier $g_n(x) = {\mathbbm 1}_{ \{ \eta_n(x) > 1/2 \}}$.

It is well-known that, in addition to this simple classifier, several other alternative
methods (kernel classifiers, neural networks, support vector machines,...) have been
developed and extensively analyzed in the latest years.
However, when used in practice with real data sets, the performance of Fisher's rule
is often found to be very close to that of the best one among all
the main alternative procedures.
On these grounds, Hand (2006) has argued in a provocative paper about the ``illusion of
progress'' in supervised classification techniques.
The central idea would be that the study of new classification rules often fails
to take into account the structure of real data sets and it tends to overlook the
fact that, in spite of the its theoretical limitations, Fisher's rule is
quite satisfactory in many practical applications.
This, together with its conceptual simplicity, explains its popularity over the years.

\

\noindent
\it 1.2 The purpose and  structure of this paper\rm

\

We are concerned here with the problem of (binary) supervised classification with
functional data. That is, we consider the general framework indicated above but
we will assume throughout that the space $({\cal F},D)$ where the random elements
$X_i$ take values is a separable metric space of functions.
For some theoretical results (Theorem 2) we will impose a more specific assumption
by taking ${\cal F}$ as the space $C[a,b]$ of real continuous functions defined in a closed
finite interval $[a,b]$, with the usual supremum norm $\Vert \; \Vert_\infty$.

The study of discrimination techniques with functional data
is not as developed as the corresponding finite-dimensional theory
but, clearly, is one of the most active research topics in
the booming field of functional data analysis (FDA).
Two well-known books including broad overviews of FDA with interesting examples are Ferraty
and Vieu (2006) and Ramsay and Silverman (2005). Other
recent more specific references will be mentioned below.

There are of course several important differences between
the theory and practice of supervised classification for
functional data and the classical development of this topic
in the finite-dimensional case, where typically the data dimension $d$ is much smaller
than the sample size $n$ (the ``high-dimensional''
case where $d$ is ``large'', and usually $d>n$, requires a
separate treatment). A first important practical
difference is the role of Fisher's linear discriminant
method as a ``default'' choice and a benchmark for
comparisons. As we have mentioned, this holds for the
finite dimensional cases with ``small'' values of $d$ but
it is not longer true if functional (or high-dimensional) data  are
involved. To begin with, there is no obvious way to apply
in practice Fisher's idea in the infinite-dimensional case,
as it requires to invert a linear operator which is not in
general a straightforward task in functional spaces; see,
however, James and Hastie (2001) for an interesting
adaptation of linear discrimination ideas to a functional
setting. Then, the question is whether there exists any functional discriminant method, based on
simple ideas, which could play a reference role similar to that of Fisher's method in the
finite dimensional case. The results in this paper suggest (as a partial, not definitive, answer)
that the $k$-NN method could  represent a ``default standard'' in functional settings.

Another difference, particularly important from
the theoretical point of view, concerns the universal
consistency of the $k$-NN classifier. A classical result by
Stone (1977) establishes that in the finite-dimensional
case (with $X_i\in{\mathbb R}^d$) the conditional error of
the $k$-NN classifier
\begin{equation} \label{CondProbErr}
L_n=P \{ g_n(X)\neq Y |\mathcal X_n\},
\end{equation}
converges in probability (and also in mean) to that of the Bayes (optimal) rule $g^*$,
that is, $E(L_n)\rightarrow L^*=P \{ g^*(X)\neq Y \}$,
provided that $k_n\to\infty$ and $k_n/n\to 0$ as
$n\to\infty$. This result
holds universally, that is, irrespective of the
distribution of the variable $(X,Y)$.
The interesting point here is that this universal
consistency result is no longer valid in the
infinite-dimensional setting. As recently proved by C\'erou
and Guyader (2006), if the space ${\cal F}$ where $X$ takes values is a
general separable metric space, a non-trivial condition must be imposed
on the distribution of $(X,Y)$ in order to ensure the
consistency of the $k$-NN classifier.

The aim of this paper is twofold, with a common focus on the $k$-NN classifier and
in close relation with the above mentioned two differences
between the classification problem in finite and infinite settings. First, on the theoretical
side, we have a further look at the consistency theorem in C\'erou and Guyader
(2006) by giving concrete non-trivial examples where their consistency condition is
fulfilled. Second, from a more practical viewpoint, we will carry out numerical
comparisons (based both on Monte Carlo studies and real data examples)
to assess the performance of different functional classifiers, including $k$-NN.

This paper is organized as follows. In Section 2 the consistency of the functional $k$-NN
classifier is established, as a consequence of Theorem 2 in
C\'erou and Guyader (2006), for a broad class of Gaussian processes.
In Section 3 other functional classifiers recently considered in the literature
are introduced and briefly commented. They are all compared through a
simulation study (based on two different models) as well as six real data examples,
very much in the spirit of Hand's (2006) paper, where the performance of the
classical Fisher's rule was assessed in terms of its discrimination  capacity in
several randomly chosen data sets.

\

\noindent
\bf 2. On the consistency of the functional $k$-NN classifier\rm

\

In the functional classification problem several auxiliary devices have been used to
overcome the extra difficulty posed by the infinite dimensional nature of the feature space.
They include dimension reduction techniques (e.g., James and Hastie 2001,
Preda {\em et al.} 2007), random projections combined with data-depth measures
projections  use of data-depth measures (Cuevas {\em et
al.} 2007) and different adaptations to the functional framework of several
non-parametric and regression-based methods, including kernel classifiers (Abraham et al. 2006, Biau et al. 2005, Ferraty and Vieu
2003), reproducing kernel procedures (Preda 2007), logistic regression (M\"uller and Stadtm\"uller 2005)
and multilayer perceptron techniques with
functional inputs (Ferr\'e and Villa 2006).

\

\noindent
\it 2.1 On the consistency of the functional $k$-NN classifier\rm

\

The functional $k$-NN classifier belongs also to the class of procedures adapted from the usual
non-parametric multivariate setup.
Nevertheless, unlike most of the above mentioned functional methodologies, the $k$-NN procedure works according to exactly the same principles
in the finite and infinite-dimensional cases. It is defined by $g_n(x) = {\mathbbm 1}_{ \{ \eta_n(x) > 1/2 \}}$,
where $\eta_n$ is the $k$-NN regression estimator (\ref{RegEstkNN}), whose definition is formally identical to that of the
finite-dimensional case.
The intuitive interpretation is also the same in both cases.
No previous data manipulation, projection or dimension
reduction technique is required in principle, apart from the
discretization process necessarily involved in the
practical handling of functional data. In the present
section we offer some concrete examples where the $k$-NN
functional classifier is weakly consistent. As we have
mentioned in the previous section, this is a non-trivial
point since the $k$-NN classifier is no longer universally consistent
in the case of infinite-dimensional inputs $X$.

Throughout this section the feature space where the
variable $X$ takes values is a separable metric space $({\cal
F},D)$. We will denote by $P_X$ the distribution of $X$
defined by
$P_X (B) = P \{  X\in B \} \quad \mbox{for } B\in\mathcal B_{\mathcal
F}$, where $\mathcal B_{\mathcal F}$ are the Borel sets of $\mathcal
F$.

Let us now consider
the following regularity assumption on the regression
function $\eta(x)=E(Y|X=x)$
\begin{description}
\item[(BC) Besicovitch condition:]
$$
\lim_{\delta\to 0} \frac{1}{P_X(B_{X,\delta})} \int_{B_{X,\delta}} \eta(z)  dP_X(z) = \eta(X)
\quad \mbox{in probability},
$$
where $B_{x,\delta} := \{ z\in \mathcal F: D(x,z)\leq \delta \}$ is the closed ball
with center $x$ and radius $\delta$.
\end{description}

Under \bf (BC)\/ \rm C\'erou and Guyader (2006, Th. 2) get the following consistency
result.

\

\noindent
\em
Denote by $L_n$ and $L^*$, respectively, the conditional error associated with the above defined $k$-NN classifier
and the Bayes (optimal) error for the problem at hand. If $({\cal F},D)$ is separable and condition \bf (BC)\/ \rm \em is
fulfilled then the $k$-NN classifier is weakly consistent,
that is $E(L_n)\rightarrow L^*$, as $n\to\infty$, provided
that $k\to\infty$ and $k/n\to 0$\rm.

\

\noindent
Besicovich condition plays an important role also in the consistency of kernel rules
(see Abraham et al. 2006).

C\'erou and Guyader (2006) have also considered the following more convenient condition (called
$P_X$-continuity) that ensures \bf (BC)\rm:
For every $\epsilon>0$ and for $P_X$-a.e. $x\in \mathcal F$
$$
\lim_{\delta\to 0} P_X \{ z\in \mathcal F: |\eta(z)-\eta(x)|>\epsilon | D(x,z)<\delta \} = 0.
$$
However, for our purposes, it will be sufficient to observe that the continuity
($P_X$-a.e.) of $\eta(x)$ implies also {\bf (BC)}.
We are interested in finding families of distributions of $(X,Y)$ under which the regression function
$\eta(x)$ is continuous ($P_X$-a.e.) and hence \bf (BC)\/ \rm holds.

From now on we will use the following notation. Let $\mu_i$ be the distribution of $X$
conditional on $Y=i$, that is,
$\mu_i(B) = P \{ X\in B|Y=i \}$, for $B\in \mathcal B_{\mathcal F}$ and $i=0,1$.
We denote by $S_i \subset \mathcal F$ the support of $\mu_i$, for $i=0,1$, and $S=S_0\cap S_1$.
The expression $\mu_0 << \mu_1$ will denote that $\mu_0$ is
absolutely continuous with respect to $\mu_1$. Also we will assume that $p=P\{Y=0\}$
fulfills $p\in(0,1)$.

The following theorem shows that the property of continuity (resp. $P_X$-continuity)
of $\eta(x)$, and hence the weak consistency of the $k$-NN classifier, follows
from the continuity (resp $P_X$-continuity) of the Radon-Nikodym derivative of $\mu_0$
with respect to $\mu_1$ provided that it exists.

\

\noindent
{\sc Theorem 1:} {\em
Assume that $P_X(\partial S)=0$ and that $\mu_0 << \mu_1$ and $\mu_1 << \mu_0$ on $S$.
Then the following inequality holds for $P_X$-a.e. $x,z\in{\cal F}$.
\begin{equation*}
|\eta(z)-\eta(x)| \leq \frac{p}{1-p} \left|\frac{d\mu_0}{d\mu_1}(x) -
    \frac{d\mu_0}{d\mu_1}(z)\right|,
\end{equation*}
where  $d\mu_0/d\mu_1$ denotes the Radon-Nikodym derivative of $\mu_0$ with respect to
$\mu_1$. When $S_0=S_1=S$ the assumption $P_X(\partial S)=0$ may be dropped.

In particular, $\eta$ is continuous $P_X$-a.e. (resp. $P_X$-continuous) whenever $d\mu_0/d\mu_1$
is continuous $P_X$-a.e. (resp. $P_X$-continuous). Of course, a similar result holds by
interchanging the sub-indices 0 and 1 and replacing $p$ by $1-p$.}

\vspace{3 mm}

\noindent
{\sc Proof:}
Define $\mu=\mu_0+\mu_1$. Then $\mu_i << \mu$,
for $i=0,1$, and we can define the Radon-Nikodym derivatives $f_i = d\mu_i/d\mu$, for $i=0,1$.
From the definition of the conditional expectation we know
that $\eta(x)=E(Y|X=x)=P(Y=1|X=x)$ can be expressed by
\begin{equation} \label{etaBayes}
\eta(x) = \frac{f_1(x)(1-p)}{f_0(x) p + f_1(x)(1-p)}.
\end{equation}
Observe that
$\mu \lvert_{S^c\cap S_i} = \mu_i\lvert_{S^c\cap S_i}$ and thus
$f_i \lvert_{S^c\cap S_i} = \mathbbm{1}_{S^c\cap S_i}$, for $i=0,1$.
Since $\mu_0 << \mu_1$ and $\mu_1 << \mu_0$ on $S$ then, on this set, we can define the
Radon-Nikodym derivatives $d\mu_0/d\mu_1$ and $d\mu_1/d\mu_0$. In this case, it also holds
that $\mu\lvert_S << \mu_i\lvert_S$, for both $i=0,1$ and
$$
\frac{d\mu}{d\mu_i}(x) = 1 + \frac{d\mu_{1-i}}{d\mu_i} (x)
\qquad \mbox{for any } x\in S.
$$
Then (see, e.g., Folland 1999), for $i=0,1$ and for $P_X$-a.e. $x\in S$,
\begin{equation} \label{DRN}
f_i(x) = \frac{d\mu_i}{d\mu}(x) = \left( \frac{d\mu}{d\mu_i}(x) \right)^{-1}
    = \frac{1}{1 + \frac{d\mu_{1-i}}{d\mu_i} (x) }
\end{equation}
Substituting (\ref{DRN}) into expression (\ref{etaBayes}) we get
\begin{eqnarray}
\eta(x) & = & \left\{ \begin{array}{l}
0 \quad \mbox{if } x\in S_0\cap S^c \\
1 \quad \mbox{if } x\in S_1\cap S^c \\
\displaystyle \frac{1-p}{p \frac{d\mu_0}{d\mu_1}(x) + 1-p} \quad \mbox{if } x\in S .
\end{array} \right.\label{etax}
\end{eqnarray}
Using this last expression we can see that if $P_X(\partial S)=0$ and if $d\mu_0/d\mu_1$
is continuous $P_X$-a.e. (resp. $P_X$-continuous) on $S$
then $\eta$ is also continuous $P_X$-a.e. (resp. $P_X$-continuous) on $S$.
To see this it suffices to observe that,
for $P_X$-a.e. $x,z\in \mbox{int}(S)$,
\begin{eqnarray*}
|\eta(z)-\eta(x)| & & = \left| \frac{1-p}{p \frac{d\mu_0}{d\mu_1}(z) + 1-p} -
\frac{1-p}{p \frac{d\mu_0}{d\mu_1}(x) + 1-p} \right| \\
    & & \leq \frac{p}{1-p} \left|\frac{d\mu_0}{d\mu_1}(x) - \frac{d\mu_0}{d\mu_1}(z)\right| .
\end{eqnarray*}
To derive the last inequality we have used that, as $\mu_i$, $i=0,1$, are positive
measures, the Radon-Nikodym derivative $d\mu_0/d\mu_1$ is also
non-negative. \hfill{$\Box$}

\

In order to be able to combine Theorem 1 and the consistency result in C\'erou and Guyader (2006, Th. 2),
we are interested in finding distributions $\mu_0,\mu_1$  of an infinite-dimensional random element $X$
such that $\mu_0 << \mu_1$ and $\mu_1 << \mu_0$ with continuous Radon-Nikodym derivatives.
Measures $\mu_0$ and $\mu_1$ satisfying that $\mu_0 << \mu_1$ and $\mu_1 << \mu_0$
on $S$ are said to be {\em equivalent} on $S$.

Let us denote by $(C[a,b],\|\;\|_\infty)$ the metric space of continuous real-valued functions $x$
defined on the interval $[a,b]$, endowed with the supremum norm,
$\| x\|_\infty=\sup\{|x(t)|:t\in [a,b]\}$. Also let  $C^{2}[a,b]$ be the space of twice
continuously differentiable functions defined on $[a,b]$.

In the next theorem we show a broad class of Gaussian processes
fulfilling the conditions of Theorem 2 in C\'erou and Guyader (2006).
Thus the consistency of the $k$-NN classifier is
guaranteed for them. A key element in the proof are the
results by Varberg (1961) and J\o rsboe (1968) providing explicit expressions for
the Radon-Nikodym derivative of a Gaussian measure with
respect to another one. From the gaussianity assumption, the model is completely
determined by giving the mean and covariance functions.
For the sake of a more clear and systematic
presentation the statement is divided into three parts: The
first one applies to the case where the mean function in
both functional populations, with distributions
$\mu_0$ and $\mu_1$ (corresponding to $X|Y=0$ and $X|Y=1$), is common and the difference between both processes lies in
the covariance functions (which however keep a common
structure). The second part considers the dual case where
the difference lies in the mean functions and the covariance structure is common. Finally, the
third part of the theorem generalizes the previous two
statements by including the case of different mean
and covariance functions.

\

\noindent
{\sc Theorem 2:} {\em
Let $(\mathcal F,D) = (C[a,b],\| \; \|_\infty)$ with $0\leq a<b<\infty$.
\begin{enumerate}[{\bf a)}]
\item Assume that $X|Y=i$, for $i=0,1$,
are Gaussian processes on $[a,b]$, whose mean function is
zero
and with covariance functions $\Gamma_i(s,t) = u_i(\min(s,t)) \, v_i(\max(s,t))$,
for $s,t\in[a,b]$, where $u_i,v_i$, for $i=0,1$, are positive functions in $C^{2}[a,b]$.
Assume also that $v_i$, for $i=0,1$, and $v_1u_1'-u_1v_1'$ are bounded away from zero on $[a,b]$,
that $u_1v_1'-u_1'v_1 = u_0v_0'-u_0'v_0$ and that $u_1(a)=0$ if and only if $u_0(a)=0$.
Then $d\mu_0/d\mu_1$ is continuous on $\mathcal F$.
\item Assume that $X|Y=i$, for $i=0,1$, are Gaussian processes on $[a,b]$,
with equal covariance function $\Gamma(s,t) = u(\min(s,t)) \, v(\max(s,t))$,
for $s,t\in[a,b]$, where $u,v\in C^{2}[a,b]$ are positive functions and
$v$ and $vu'-uv'$ are bounded away from zero on $[a,b]$.
Assume also that the mean function of $X|Y=1$ is 0 and that of $X|Y=0$ is a function $m\in C^2[a,b]$,
such that $m(a)=0$ whenever $u(a)=0$. Then $d\mu_0/d\mu_1$ is continuous on $\mathcal F$.
\item Assume that $X|Y=i$, for $i=0,1$, are Gaussian processes on $[a,b]$,
with mean functions $m_i\in C^{2}[a,b]$
and covariance functions $\Gamma_i(s,t) = u_i(\min(s,t)) \, v_i(\max(s,t))$,
for $s,t\in[a,b]$, where $u_i,v_i$, for $i=0,1$, are positive functions in $C^{2}[a,b]$
which fulfill the same conditions imposed in (a).
Assume also that $m_i(a)=0$ whenever $u_i(a)=0$.
Then $d\mu_0/d\mu_1$ is continuous on $\mathcal F$.
\end{enumerate}
Therefore, under the assumptions in either (a), (b) or (c), the $k$-NN classifier
discriminating between $\mu_0$ and $\mu_1$ is weakly consistent when $k\to\infty$ and $k/n\to 0$.
}

\

\noindent
{\sc Proof:} \begin{enumerate}[{\bf a)}]
\item Varberg (1961, Th. 1) shows that, under the assumptions of (a),
$\mu_0$ and $\mu_1$ are equivalent measures and the Radon-Nikodym derivative of
$\mu_0$ with respect to $\mu_1$ is given by
\begin{equation} \label{J1}
\frac{d\mu_0}{d\mu_1}(x) = C_1 \, \exp\left\{ \frac{1}{2} \left[ C_2 x^2(a) +
\int_a^b f(t) d\left( \frac{x^2(t)}{v_0(t)v_1(t)} \right) \right] \right\}
\end{equation}
where
$$
C_1 = \left\{ \begin{array}{l}
\left( \frac{v_0(a)v_1(b)}{v_0(b)v_1(a)} \right)^{1/2} \quad \mbox{if } u_0(a)=0 \\
\left( \frac{u_1(a)v_1(b)}{v_0(b)u_0(a)} \right)^{1/2} \quad \mbox{if } u_0(a)\ne 0
\end{array} \right.
\qquad
C_2 = \left\{ \begin{array}{l}
0 \quad \mbox{if } u_0(a)=0 \\
\left( \frac{v_0(a)u_0(a)-u_1(a)v_1(a)}{v_1(a)v_0(a)u_0(a)u_1(a)} \right)^{1/2} \quad \mbox{if } u_0(a)\ne 0
\end{array} \right.
$$
and
$$
f(s) = \frac{v_1(s)v_0'(s)-v_0(s)v_1'(s)}{v_1(s)u_1'(s)-u_1(s)v_1'(s)} \quad \mbox{for } s\in [a,b] .
$$
Observe that, by the assumptions of the theorem, this function $f$ is
differentiable with bounded derivative.
Thus $f$ is of bounded variation and it may be expressed as the difference
of two bounded positive increasing functions.
Therefore the stochastic integral (\ref{J1}) is
well defined and it can be evaluated integrating by parts,
$$
\frac{d\mu_0}{d\mu_1}(x) = C_1 \exp \left[ \frac{1}{2} \left( C_3x^2(a) + C_4 x^2(b)
 - \int_a^b \frac{x^2(t)}{v_0(t)v_1(t)} df(t)\right) \right]
$$
with
$ C_3=C_2-f(a)/v_0(a)v_1(a) $ and $ C_4 = f(b)/v_0(b)v_1(b) $.
It is clear that this derivative is a continuous functional of $x$ with respect
to the supremum norm.

Now, Theorem 1 implies that $\eta(x)$ is continuous and, therefore, Besicovich condition
{\bf (BC)} holds and, from Theorem 2 in C\'erou and Guyader (2006), the $k$-NN classifier
is weakly consistent. Note that the equivalence of $\mu_0$ and $\mu_1$ implies
the coincidence of both supports $S_0=S_1=S$.

\item In J\o rsboe (1968), p. 61, it is proved that, under the indicated assumptions, $\mu_0$ and $\mu_1$ are
equivalent measures with the following Radon-Nikodym derivative
$$
\frac{d\mu_0}{d\mu_1}(x) = \exp \left\{ D_1 + D_2 \, x(a) + \frac{1}{2} \int_a^b g(t)
d\left( \frac{2x(t)-m(t)}{v(t)} \right) \right\}
$$
where
$$
D_1 = -\frac{m^2(a)}{2 \, u(a) \, v(a)} \mathbbm 1_{\{ u(a)>0 \}} \; , \qquad
D_2 = \frac{m(a)}{u(a) \, v(a)} \mathbbm 1_{\{ u(a)>0 \}}
$$
and
$$
g(t) = \frac{v(t)m'(t)-m(t)v'(t)}{v(t)u'(t)-u(t)v'(t)} \; .
$$
Again, the integration  by parts gives
\begin{equation}
\frac{d\mu_0}{d\mu_1}(x) = \exp \left\{ D_3 + \left( D_2 -2\,\frac{g(a)}{v(a)} \right) x(a)
    + 2 \,\frac{g(b)}{v(b)}\, x(b) - 2 \int_a^b \frac{x(t)}{v(t)}\, dg(t) \right\} ,
\end{equation}
with
$$
D_3 = D_1 - \int_a^b g(t) \, d\left( \frac{m(t)}{v(t)} \right) .
$$
Thus $d\mu_0/d\mu_1$, and hence $\eta$, are continuous and the consistency of the
$k$-NN classifier holds also in this case.
\item Let us denote by $P_{m,\Gamma}$ the distribution of the Gaussian process with mean $m$
and covariance function $\Gamma$. Then $\frac{d\mu_0}{d\mu_1}(x)$ is continuous since
(see e.g. Folland 1991)
\begin{equation} \label{RNNS}
\frac{d\mu_0}{d\mu_1}(x) = \frac{dP_{m_0,\Gamma_0}}{dP_{m_1,\Gamma_1}} (x)
    = \frac{dP_{m_0,\Gamma_0}}{dP_{0,\Gamma_0}} (x) \,
      \frac{dP_{0,\Gamma_0}}{dP_{0,\Gamma_1}} (x) \,
      \frac{dP_{0,\Gamma_1}}{dP_{m_1,\Gamma_1}} (x),
\end{equation}
and, as we have shown in the proofs of (a) and (b), the Radon-Nikodym derivatives
in the right-hand side of (\ref{RNNS}) are all continuous. \hfill{$\Box$}
\end{enumerate}

\

\noindent
{\sc Remark 1 (Application to the Ornstein-Uhlenbeck processes).}
Let $X|Y=i$, for $i=0,1$, be Gaussian processes on $[a,b]$, with zero mean
and covariance function $\Gamma_i(s,t) = \sigma_i^2 \exp(-\beta_i|s-t|)$, for $s,t\in[a,b]$,
where $\beta_i,\sigma_i>0$ for $i=0,1$.
Assume that $\sigma_1^2\beta_1=\sigma_0^2\beta_0$.
Then these processes satisfy the assumptions in Theorem 2(a).
\vspace{3 mm}

\noindent
{\sc Remark 2 (Application to the Brownian motion).}
Theorem 2(b) can also be used to consistently discriminate between a
Brownian motion without trend ($m_0=0$) and another one
with trend ($m_1\neq 0$). It will suffice to consider the
case where $u(t)=t$ and $v\equiv 1$.
\vspace{3 mm}

\noindent
{\sc Remark 3 (On triangular covariance functions).}
Covariance functions of type $\Gamma(s,t) = u(\min(s,t)) \, v(\max(s,t))$,
called \it triangular\rm, have received considerable attention in the literature.
For example, Sacks and Ylvisaker (1966) use this condition in the study of optimal
designs for regression problems where the errors are generated by a zero
mean process with covariance function $K(s,t)$. It turns out that the Hilbert space
with reproducing kernel $K$ plays an important role in the results and, as these authors
point out, the norm of this space is particularly easy to handle when $K$ is triangular.
On the other hand, Varberg (1964) has given an interesting representation of the
processes $X(t),\ 0\leq t<b$, with zero mean and triangular covariance function by
proving that they can be expressed in the form
$$
X(t)=\int_0^bW(u)d_uR(t,u),
$$
where $W$ is the standard Wiener process and $R=R(t,u)$ is a function, of bounded
variation with respect to $u$, defined in terms of $K$.
\vspace{3 mm}

\noindent
{\sc Remark 4 (On plug-in functional classifiers).} The explicit knowledge of the conditional
expectation (\ref{etax}) in the cases considered in Theorem 2 could be
explored from the statistical point of view as they suggest
to use ``plug-in'' classifiers obtained by replacing
$\eta(x)$ in (\ref{opt}) with suitable parametric or semiparametric estimators.
\vspace{3 mm}

\noindent
{\sc Remark 5 (On equivalent Gaussian measures and their supports).}
According to a well-known result by Feldman and H\'ajek,
for any given pair of Gaussian processes, there is a dichotomy in such
a way that they are either equivalent or mutually singular.
In the first case both measures $\mu_0$ and $\mu_1$ have a common support $S$ so that
Theorem 1 is applicable with $S=S_0=S_1$. As for the identification of the support,
Vakhania (1975) has proved that if a Gaussian process, with trajectories in a
separable Banach space ${\cal F}$, is not degenerate
(i.e., then the distribution of any non-trivial linear continuous functional is not degenerate)
then the support of such process is the whole space ${\cal F}$. Again, expression (\ref{etax}) of
the regression functional $\eta$ suggests the possibility of investigating
possible nonparametric estimators for the Radon-Nikodym derivative $d\mu_0/d\mu_1$ which would
in turn provide plug-in versions of the Bayes rule $g^*(x) = {\mathbbm 1}_{ \{ \eta(x) > 1/2 \}}$
with no further assumption on the structure of the involved Gaussian processes, apart from
their equivalence.

\

\noindent
\bf 3. Some numerical comparisons\rm

\

The aim of this section is to compare (numerically) the performance of several
supervised functional classification procedures already introduced in
the literature. The procedures are the $k$-NN rule, computed both with respect to the
supremum norm $\|\;\|_\infty$ and the $L^2$ norm $\|\;\|_2$, and other discrimination
rules reviewed in Section 3.1. One of the objectives of this numerical study is
to have some insight into which classification procedures perform
well no matter the type of functional data under consideration and could thus
be considered a sort of benchmark for the functional discrimination problem.
Section 3.2 contains a Monte Carlo study carried out on two different functional
data generating models. In Section 3.3 we consider six functional real data
sets taken from the literature.

\

\noindent
\it 3.1 Other functional classifiers\rm

\

Here we will review other classification techniques that have been used
in the literature in the context of functional data.
From now on we denote by $(t_1,\ldots,t_N)$
the nodes where the functional predictor $X$ has been
observed.
\vspace{5 mm}

\noindent
{\em Partial Least Squares (PLS) classification} \vspace{3 mm}

Let us first describe the procedure in the context of a multivariate predictor $\mathbf X$.
PLS is actually a dimension reduction technique for regression problems with predictor
$\mathbf X$ and a response $Y$ (which in the case of classification takes only two values, 0 or 1,
depending on which population the individual comes from). The dimension reduction is carried out by projecting
$\mathbf X$ onto an lower dimensional space such that the
coordinates of the projected $\mathbf X$,
the PLS coordinates, are uncorrelated to each other and have maximum covariance with $Y$.
Then, if the aim is classification, Fisher's linear discriminant is applied to the PLS
coordinates of $\mathbf X$ (see Barker and Rayens 2003, Liu and Rayens 2007).
In the case of a functional predictor $X$ (see Preda et al. 2007), the above described procedure is applied to
the discretized version of $X$, $\mathbf X=(X(t_1),X(t_2),\ldots,X(t_N))$.
Here we have chosen the number of PLS directions, among the values 1,\ldots,10, by cross-validation.
\vspace{3 mm}

\noindent
{\em Reproducing Kernel Hilbert Space (RKHS) classification} \vspace{3 mm}

We will also define this technique initially for a multivariate predictor $\mathbf X$.
For simplicity, we will assume that $\mathbf X$ takes values in $[0,1]^N$.
Let $\kappa$ be a function defined on $[0,1]^N\times[0,1]^N$.
A RKHS with kernel $\kappa$ is the vector space generated by all finite linear combinations
of functions of the form $\kappa_{\mathbf t^*}(\cdot)=\kappa(\mathbf t^*,\cdot)$,
for any $\mathbf t^*\in[0,1]^N$, and endowed with the inner product given by
$\langle \kappa_{\mathbf t^*}, \kappa_{\mathbf t^{**}}\rangle_\kappa=\kappa(\mathbf t^*,\mathbf t^{**})$.
RKHS are frequently used in the context of Machine Learning
(see Evgeniou {\em et al.} 2002, Wahba 2002); for their applications in Statistics the reader
is referred to the monograph of Berlinet and Thomas-Agnan (2004).
In this work we use the Gaussian kernel $\kappa(\mathbf s,\mathbf t) = \exp( -\|\mathbf s-\mathbf t\|_2^2/\sigma_\kappa^2 )$, where
$\sigma_\kappa>0$ is a fixed parameter. The classification problem is solved by plugging a regression
estimator of the type $\eta_n(\mathbf x) = \sum_{i=1}^n c_i \, \kappa(\mathbf x,\mathbf X_i)$ into
the Bayes classifier. When $X$ is a random function, this procedure is applied
to the discretized $X$.
The parameters $c_i$, for $i=1,\ldots,n$, are chosen to minimize the risk functional
$n^{-1} \sum_{i=1}^n (Y_i-\eta_n(X_i))^2 + \lambda \langle \eta,\eta\rangle_\kappa$,
where $\lambda>0$ is a penalization parameter.
In this work the values of the parameters $\lambda$ and $\sigma_\kappa$ have been chosen
by cross-validation via a leave-one-out procedure.
According to our results, it seems that the performance the RKHS methodology
is rather sensitive to changes in these parameters and even to the starting point of the
leave-one-out procedure mentioned. \vspace{3 mm}

\noindent
{\em Classification via depth measures} \vspace{3 mm}

The idea is to assign a new observation $x$ to that population, $P_0$ or $P_1$, with
respect to which $x$ is deeper (see Ghosh and Chaudhuri 2005, Cuevas et al. 2007).
From the five functional depth measures considered by Cuevas et al. (2007) we have
taken the $h$-mode depth and the random projection (RP) depth.

Specifically, the $h$-mode depth of $x$ with respect to the population given by the random
element $X$ is defined as $f_h(x) = E(K_h(\|x-X\|_2))$, where $K_h(\cdot) = h^{-1} K(\cdot/h)$,
$K$ is a kernel function (here we have taken the Gaussian kernel
$K(t) = \sqrt{2/\pi} \exp(-t^2/2)$) and $h$ is a smoothing parameter.
As the distribution of $X$ is usually unknown, in the simulations we actually use
the empirical version of $f_h$,
$ \hat f_h(x) = n^{-1} \sum_{i=1}^n K_h(\|x-X_i\|_2) $.
The smoothing parameter has been chosen as the 20 percentile in the $L^2$ distances between
the functions in the training sample (see Cuevas et al. 2007).

To compute the RP depth the training sample $X_1,\ldots,X_n$
is projected onto a (functional) random direction $a$ (independent of the $X_i$).
The sample depth of an observation $x$ with respect to $P_i$ is defined
as the univariate depth of
the projection of $x$ onto $a$ with respect to the projected training sample from $P_i$.
Since $a$ is a random element this definition leads
to a random measure of depth, but a single representative value has been obtained
by averaging these random depths over 50 independent random directions
(see Cuevas and Fraiman 2008 for a certain theoretical development of this idea).
If we are working with discretized versions $(x(t_1),\ldots,x(t_N))$ of the
functional data $x(t)$, we may take $a$ according to a uniform distribution on the
unit sphere of ${\mathbb R}^N$. This can be achieved, for example, setting
$a=Z/\|Z\|$, where $Z$ is drawn from standard Gaussian distribution on ${\mathbb R}^N$.
\vspace{3 mm}

\noindent
{\em Moving window rule} \vspace{3 mm}

The moving window classifier is given by
$$
g_n(x) = \left\{ \begin{array}{ll}
0 & \mbox{if } \sum_{i=1}^n \mathbbm{1}_{\{Y_i=0,X_i\in B(x,h)\}}
\geq \sum_{i=1}^n \mathbbm{1}_{\{Y_i=1,X_i\in B(x,h)\}}  , \\
1 & \mbox{otherwise} ,
\end{array} \right.
$$
where $h=h_n>0$ is a smoothing parameter. This classification rule was considered
in the functional setting, for instance, by Abraham et al. (2006). In this work the
parameter $h$ has been chosen again via cross-validation.

\

\noindent
\it 3.2 Monte Carlo results\rm

\

In this section we study two functional data models already considered by other authors.
More specifically, in Model 1, similar to one used in Cuevas et al.
(2007), $X|Y=i$ is a Gaussian process with mean
$ m_i(t) = 30 \, (1-t)^{1.1^i} \, t^{1.1^{1-i}} $
and covariance function $\Gamma_i(s,t)=0.25\exp(-|s-t|/0.3)$, for $i=0,1$.
Observe that this model with smooth trajectories satisfies the assumptions in
Theorem 2 and thus we would expect the $k$-NN classification rule
(with respect to the $\|\;\|_\infty$ norm) to perform nicely. Let us note that the value
of 1.1 in the exponent of $m_i(t)$ is in fact the one used in Model 1, pg. 487, of Cuevas et al.
(2007), although in their work a 1.2 was misprinted instead.

Model 2 appears in Preda et al. (2007), but here the
functions $h_i$, used to define the mean, have been rescaled to have domain
$[0,1]$. The trajectories of $X|Y=i$ are given by
\begin{equation} \label{Model2}
X_i(t)=U \, h_1(t) + (1-U) \, h_{i+2}(t) + \epsilon(t) \qquad \mbox{for } i=0,1,
\end{equation}
where $U$ is uniformly distributed on $[0,1]$, $h_1(t) = 2 \max(3-5|2t-1|,0)$,
$h_2(t) = h_1(t-1/5)$, $h_3(t) = h_1(t+1/5)$ and the $\epsilon(t)$ is
an approximation to the continuous-time white noise.
In practice, this means that in the discretized approximations $(X(t_1),\ldots,X(t_N))$
to $X(t)$, the variables $\epsilon(t_1),\ldots,\epsilon(t_N)$ are independently drawn
from a standard normal distribution.

The simulation results are summarized in Tables 1 and 2.
The number of equispaced nodes where the functional data have been evaluated is
the same for both models, $51$.
The number of Monte Carlo runs is 100.
In every run we generated two training samples (from $X|Y=0$
and $X|Y=1$ respectively) each with sample size
100, and we also generated a test sample of size 50 from each of the two
populations.  The tables display the descriptive statistics of the proportion of
correctly classified observations from these test samples.

\

\begin{table}[h] \small \label{SimMod1}
\begin{center}
\begin{tabular}{rcccccccc}
                   & $k$-NN$|_\infty$ & $k$-NN$|_2$ & PLS & RKHS & $h$-modal & RP(hM) & MWR \\ \hline
Minimum            & 0.6200 & 0.6600 & 0.6000 & 0.4800 & 0.6400 & 0.5400 & 0.6600 \\
First quartile     & 0.8000 & 0.8000 & 0.8000 & 0.6600 & 0.8000 & 0.7800 & 0.8000 \\
Median             & 0.8400 & 0.8400 & 0.8400 & 0.8400 & 0.8400 & 0.8400 & 0.8400 \\
Mean               & 0.8396 & 0.8354 & 0.8371 & 0.7999 & 0.8409 & 0.8260 & 0.8393 \\
Third quartile     & 0.8800 & 0.8800 & 0.8800 & 0.9400 & 0.8800 & 0.8800 & 0.8800 \\
Maximum            & 0.9800 & 0.9600 & 0.9800 & 1.0000 & 0.9800 & 0.9800 & 1.0000 \\ [2 mm]
Std. deviation     & 0.0603 & 0.0572 & 0.0668 & 0.1457 & 0.0589 & 0.0725 & 0.0634 \\ \hline
\end{tabular}
\end{center}
\caption{Simulation results for Model 1}
\end{table}

\begin{table}[h] \small \label{SimMod2}
\begin{center}
\begin{tabular}{rccccccccc}
                   & $k$-NN$|_\infty$ & $k$-NN$|_2$ & PLS & RKHS & $h$-modal & RP(hM) & MWR \\ \hline
Minimum            & 0.8400 & 0.8400 & 0.8800 & 0.8400 & 0.8600 & 0.8400 & 0.8200 \\
First quartile     & 0.9200 & 0.9400 & 0.9600 & 0.9600 & 0.9400 & 0.9400 & 0.9400 \\
Median             & 0.9600 & 0.9600 & 0.9800 & 0.9800 & 0.9800 & 0.9600 & 0.9600 \\
Mean               & 0.9522 & 0.9558 & 0.9686 & 0.9688 & 0.9657 & 0.9522 & 0.9570 \\
Third quartile     & 0.9800 & 0.9800 & 0.9800 & 1.0000 & 1.0000 & 0.9800 & 0.9800 \\
Maximum            & 1.0000 & 1.0000 & 1.0000 & 1.0000 & 1.0000 & 1.0000 & 1.0000 \\ [2 mm]
Std. deviation     & 0.0335 & 0.0355 & 0.0279 & 0.0313 & 0.0308 & 0.0345 & 0.0349 \\ \hline
\end{tabular}
\caption{Simulation results for Model 2}
\end{center}
\end{table}

Regarding Model 1, observe that there is little difference between the correct classification rates
of any of the methods, except for the RKHS procedure which performs worse. In Model 2
the PLS, RKHS and $h$-modal methods slightly outperform the others. When the Monte Carlo
study with this model was carried out, we also applied the $k$-NN classification procedures
to a spline-smoothed version of the $X$ trajectories. The result was that the mean correct
classification rate increased to 0.9582 in the case of the supremum norm and to 0.9624
in the case of the $L^2$ norm. This, together with the analysis of the flies data in the
next subsection, seems to suggest that, when the curves $X$ are irregular, smoothing these
functions will enhance the $k$-NN discrimination procedure.

\

\noindent
\it 3.3. Some comparisons based on real data sets\rm

\

\noindent
\it 3.3.1. Brief description of the data sets \rm

\

\noindent
{\em Berkeley Growth Data:}
The Berkeley Growth Study (Tuddenham and Snyder 1954) recorded the heights of
$n_0=54$ girls and $n_1=39$ boys between the ages of 1 and 18 years. Heights were measured
at 31 ages for each child. These data have been previously analyzed
by Ramsay and Silverman (2002).
\vspace{3 mm}

\noindent
{\em ECG data:} These are electrocardiogram (ECG) data, studied by Wei and Keogh (2006),
from the MIT-BIH Arrhythmia database (see Goldberger {\em et al.} 2000).
Each observation contains the successive measurements recorded
by one electrode during one heartbeat and was normalized and rescaled to have length 85.
A group of cardiologists have assigned a label of normal or abnormal to each data record.
Due to computational limitations, of the original $2026$ records in the data set,
we have randomly chosen only $200$ observations from each group.
\vspace{3 mm}

\noindent
{\em MCO data:} The variable under study is the mitochondrial calcium overload (MCO), measured
every 10 seconds during an hour in isolated mouse cardiac cells.
The data come from research conducted by Dr. David Garc\'{\i}a-Dorado at the Vall d'Hebron
Hospital (see Ruiz-Meana et al. 2003, Cuevas, Febrero and Fraiman 2004, 2007).
In order to assess if a certain drug increased the MCO level,
a sample of functions of size $n_0 = 45$ was taken from a control group
and $n_1 = 44$ functions were sampled from the treatment group.
\vspace{3 mm}

\noindent
{\em Spectrometric data:} For each of 215 pieces of meat a spectrometer provided the absorbance attained at 100
different wavelengths (see Ferraty and Vieu 2006 and references therein). The fat content of the meat was also obtained via chemical processing
and each of the meat pieces was classified as low- or high-fat.
\vspace{3 mm}

\noindent
{\em Phoneme data:} The $X$ variable is the log-periodogram (discretized to 150 nodes) of a phoneme. The
two populations correspond to phonemes ``aa'' and ``ao'' respectively (see more information in
Ferraty and Vieu 2006).
We have considered a sample of 100 observations from each phoneme.
\vspace{3 mm}

\noindent
{\em Medflies data:} This dataset was obtained by Prof. Carey from U.C. Davis (see Carey et al. 1998)
and has been studied, for instance, by M{\"u}ller and Stadtm{\"u}ller (2005).
The predictor $X$ is the number of eggs laid daily by a Mediterranean fruit fly for
a 30-day period. The fly is classified as long-lived if its remaining lifetime past 30 days
is more than 14 days and short-lived otherwise. The number of long- and short-lived flies
observed was 256 and 278 respectively.

\

\noindent
\it 3.3.2. Results \rm

\

We have applied the classification techniques reviewed in Section 3.1 to the real data sets
just described. While carrying out the simulations of Subsection 3.1,
we observed that the performance of the RKHS procedure was
very dependent on the initial values of the parameters $\sigma_K$ and $\lambda$ provided for
the cross-validation algorithm. In fact, finding initial values
for these parameters that would finally yield competitive results with respect to the
other methods took a considerable time. Thus we decided to exclude the RKHS classification
method from the study with real data.

We have computed, via a cross-validation procedure, the mean correct classification rates
attained by the different discrimination methods on the real data sets.
In Table 3 we display the results.
Since the egg-laying trajectories in the medflies data set were very irregular and spiky,
we have computed the correct classification rate for both the original data and
a smoothed version obtained with splines. The smoothing leads to a better
performance of the $k$-NN procedure with the supremum metric, just as it happened in the
simulations with Model 2.

\begin{table}[h] \label{RealDat}
\begin{center}
\begin{tabular}{lcccccccc}
Data set         & $k$-NN$|_\infty$ & $k$-NN$|_2$ & PLS    & $h$-modal & RP(hM) & MWR    \\ \hline
Growth                & 0.9462           & 0.9677      & 0.9462 & 0.9462    & 0.9462 & 0.9570 \\
ECG                & 0.9900           & 0.9950      & 0.9825 & 0.9900    & 0.8575 & 0.8850 \\
MCO                & 0.8427           & 0.8315      & 0.8876 & 0.7640    & 0.7079 & 0.6854 \\
Spectrometric                & 0.9070           & 0.8558      & 0.9163 & 0.6791    & 0.6930 & 0.6558 \\
Phoneme                & 0.7300           & 0.7800      & 0.7400 & 0.7300    & 0.7450 & 0.6950 \\
Medflies (non-smoothed) & 0.5468           & 0.5412      & 0.5262 & 0.4925    & 0.5056 & 0.5431 \\
\hspace{15 mm}
  (smoothed)     & 0.5712           & 0.5431      & 0.5094 & 0.5075    & 0.5543 & 0.5206 \\ \hline
\end{tabular}
\end{center}
\caption{Mean correct classification rates for the real data sets}
\end{table}

As a conclusion we would say that the $k$-NN classification methodology with respect to the
$L^\infty$ norm is always among the best performing ones if the $X$ trajectories are smooth.
The $k$-NN procedure with respect to the $L^2$ norm and the PLS methodology give also
good results, although the latter has the drawback of a much higher computation time.

\

\begin{center}
\sc References\rm
\end{center}

\begin{list}{}{\leftmargin .5cm\listparindent -.5cm}
\item\hspace{-.2cm}

\vspace{-.3cm}

Abraham, C., Biau, G. and Cadre, B. (2006). On the kernel rule for function classification.
Annals of the Institute of Statistical Mathematics 58, 619-633.

Barker M. and Rayens W. (2003). Partial least squares for discrimination. Journal of
Chemometrics 17, 166-73.

Berlinet, A. and Thomas-Agnan, C. (2004). Reproducing Kernel Hilbert Spaces in
Probability and Statistics. Kluwer Academic Publishers.

Biau, G., Bunea, F. and Wegkamp, M. (2005). Functional classification in Hilbert spaces. IEEE
Transactions on Information Theory 51, 2163-2172.

Carey, J.R., Liedo, P., M{\"u}ller, H.G., Wang, J.L. and Chiou, J.M. (1998).
Relationship of age patterns of fecundity to mortality, longevity, and lifetime
reproduction in a large cohort of Mediterranean fruit fly females. Journal of
Gerontology, Ser. A 53, 245--251.

C\'erou, F. and Guyader, A. (2006). Nearest neighbor classification in infinite dimension.
ESAIM: Probability and Statistics 10, 340-355.

Cuevas, A., Febrero, M and Fraiman, R. (2004). An ANOVA test for functional data.
Computational Statistics and Data Analysis 47, 111--122.

Cuevas, A., Febrero, M and Fraiman, R. (2007). Robust estimation and
classification for functional data via projection-based depth notions.
Computational Statistics 22, 481--496.

Cuevas, A. and Fraiman, R. (2008). On depth measures and dual statistics.
A methodology for dealing with general data. \it
Manuscript\rm.

Devroye, L., Gy\"orfi, L. and Lugosi, G. (1996).  A Probabilistic Theory of
Pattern Recognition. Springer-Verlag.

Evgeniou , T., Poggio, T. Pontil, M. and Verri, A. (2002). Regularization
and statistical learning theory for data analysis. Computational Statistics
and Data Analysis, 38, 421--432.

Ferraty, F. and Vieu, P. (2003). Curves discrimination: A nonparametric functional approach.
Computational Statistics and Data Analysis 44, 161--173.

Ferraty, F. and Vieu, P. (2006). Nonparametric Modelling for Functional Data. Springer.

Ferr\'e, L. and Villa, N. (2006). Multilayer perceptron with functional inputs:
an inverse regression approach. Scandinavian Journal of Statistics 33, 807--823,

Fisher, R.A. (1936). The use of multiple measurements in taxonomic problems.
Annals of Eugenics 7, 179--188.

Folland, G. B. (1999). Real analysis. Modern techniques and their applications. Wiley.

Ghosh, A. K. and Chaudhuri, P. (2005). On maximal depth and related classifiers.
Scandinavian Journal of Statistics 32, 327--350.

Goldberger, A., Amaral, L., Glass, L., Hausdorff, J., Ivanov, P.,
Mark, R., Mietus, J., Moody, G., Peng, C., and He, S. (2000).
PhysioBank, PhysioToolkit, and PhysioNet: Components of a
New Research Resource for Complex Physiologic Signals. Circulation 101, 215--220.

Hand, D.J. (1997). Construction and Assessment of Classification Rules. Wiley.

Hand, D.J. (2006). Classifier technology and the illusion of progress.
Statistical Science 21, 1--14.

Hastie, T., Tibshirani, R. and Friedman, J. (2001). The Elements of Statistical
Learning. Springer.

James, G.M. and Hastie, T.J. (2001). Functional linear discriminant analysis for
irregularly sampled curves. Journal of the Royal Statistical Society, Ser. B 63, 533-550.

J\o rsboe, O. G. (1968). Equivalence or Singularity of Gaussian Measures on
Function Spaces. Various Publications Series, No. 4, Matematisk Institut, Aarhus
Universitet, Aarhus.

Liu, Y. and Rayens, W. (2007). PLS and dimension reduction for
classification. Computational Statistics 22, 189--208.

M\"uller, H.G.  and Stadtm\"uller, U. (2005). Generalized functional linear models.
The Annals of Statistics 33, 774-805.

Preda, C. (2007). Regression models for functional data by reproducing kernel
Hilbert spaces methods. Journal of Statistical Planning and Inference 137,
829--840.

Preda, C., Saporta, G. and L\'ev\'eder, C. (2007). PLS classification of functional
data. Computational Statistics 22, 223--235.

Ramsay, J.O. and Silverman, B.W. (2002). Applied Functional Data
Analysis. Methods and Case Studies. Springer-Verlag.

Ramsay, J.O. and Silverman, B.W. (2005). Functional Data Analysis. Second edition. Springer.

Ruiz-Meana, M., Garc\'{\i}a-Dorado, D., Pina, P., Inserte, J., Agull\'o, L. and Soler-Soler, J. (2003). Cariporide preserves mitochondrial proton gradient and
delays ATP depletion in cardiomyocites during ischemic conditions.
American Journal of Physiology - Heart and Circulatory Physiology 285, 999--1006.

Sacks, J. and Ylvisaker, N.D. (1966). Designs for regression problems with correlated errors.
Annals of Mathematical Statistics 37, 66--89.

Stone, C. J. (1977). Consistent nonparametric regression. The Annals of  Statistics 5, 595-645.

Tuddenham, R. D. and Snyder, M. M. (1954). Physical growth of California boys
and girls from birth to eighteen  years. University of California Publications
in Child Development 1, 183--364.

Vakhania, N.N. (1975). The topological support of Gaussian measure in Banach space.
Nagoya Mathematical Journal 57, 59--63.

Varberg, D.E. (1961). On equivalence of Gaussian measures.
Pacific Journal of Mathematics 11, 751--762.

Varberg, D.E. (1964). On Gaussian measures equivalent to Wiener measure. Transactions
of the American Mathematical Society 113, 262--273.

Wahba, G. (2002). Soft and hard classification by reproducing kernel Hilbert space methods.
Proceedings of National Academy of Sciences 99, 16524--16530.

Wei, L. and Keogh, E. (2006). Semi-Supervised Time Series Classification.
Proceedings of the 12th ACM SIGKDD International Conference on Knowledge Discovery
and Data Mining, 748--753, Philadelphia, U.S.A.

\end{list}

\end{document}